\documentclass{article}

\newcommand{\K}{{\mathcal K}}
\newcommand{\MB}[1]{\mathbf{#1}}
\newcommand{\MBB}[1]{\mathbb{#1}}
\newcommand{\MC}[1]{\mathcal{#1}}

\newcommand{\Tr}[0]{\mathrm{Tr}}

\newcommand{\smalleqb}[1]
{
	\begingroup
	\makeatletter
	\def
	\f@size{1}
	{#1}
    \endgroup
}

\usepackage{adjustbox}
\usepackage{changepage}
\usepackage[hidelinks]{hyperref}
\usepackage{amsmath,amssymb}
\usepackage{multirow}
\usepackage{subcaption}
\usepackage[linesnumbered,boxed]{algorithm2e}
\usepackage{graphicx}
\usepackage{epstopdf}
\usepackage{url}

\hypersetup{
	pdfstartview	= {FitH},
	pdftitle			= {Non-Negative Local Sparse Coding for Subspace Clustering},
	pdfsubject		= {IDA 2018},
	pdfauthor 		= {Babak Hosseini and Barbara Hammer},
	pdfcreator 		= {XeLaTeX},
	pdfproducer		= {LaTeX with hyperref}
}

\newtheorem{proposition}{Proposition}
\newtheorem{definition}{Definition}
\newtheorem{proof}{proof}

\SetKwInput{KwInput}{Input}
\SetKwInput{KwParameters}{Parameters}
\SetKwInput{KwOutput}{Output}
\SetKwInput{KwTask}{Task}
\SetKwInput{KwInit}{Initialization}
\SetKwInput{KwProc}{Loop}
\SetKwInput{KwStepk}{Step K}
\SetKwInput{KwPre}{Pre-calculations}
\SetKwInput{Kwinit}{Initialization}
\SetKwInput{Kwhidenum}{}

\DeclareMathOperator*{\argmin}{arg\,min}

\newcommand{\ny}{{\vec{x}}}
\newcommand{\nY}{\mathbf{X}}
\newcommand{\nxs}{\gamma}
\newcommand{\nx}{{\vec{\gamma}}}
\newcommand{\nX}{\mathbf{\Gamma}}

\newcommand{\nT}{T_0}
\newcommand{\nh}{{\vec{l}}}

\newcommand{\DW}[3]{\MC{D}(\MB{#1}_{#2},\nX^{#3})}

\date{\footnotesize{Preprint of the publication~\cite{hosseini2018nonida}, as provided by the authors.\\
The final publication is available at Springer via \url{http://dx.doi.org/10.1007/978-3-030-01768-2_12} }}
\begin{document}


\title{Non-Negative Local Sparse Coding for\\ Subspace Clustering}

\author{Babak Hosseini and Barbara Hammer
\vspace{.2cm}\\
CITEC centre of excellence,
Bielefeld University, Germany\\
emails:\{bhosseini, bhammer\}@techfak.uni-bielefeld.de}

\maketitle

\begin{abstract}
	Subspace sparse coding (SSC) algorithms have proven to be beneficial to clustering problems. They provide an alternative data representation in which the underlying structure of the clusters can be better captured. However, most of the research in this area is mainly focused on enhancing the sparse coding part of the problem. 
	In contrast, we introduce a novel objective term in our proposed SSC framework which focuses on the separability of data points in the coding space.         
	We also provide mathematical insights into how this local-separability term improves the clustering result of the SSC framework.
	Our proposed non-linear local SSC algorithm (NLSSC) also benefits from the efficient choice of its sparsity terms and constraints.
	The NLSSC algorithm is also formulated in the kernel-based framework (NLKSSC) which can represent the nonlinear structure of data.        
	In addition, we address the possibility of having redundancies in sparse coding results and its negative effect on graph-based clustering problems. We introduce the link-restore post-processing step to improve the representation graph of non-negative SSC algorithms such as ours.                
	Empirical evaluations on well-known clustering benchmarks show that our proposed NLSSC framework results in better clusterings compared to the state-of-the-art baselines and demonstrate the effectiveness of the link-restore post-processing in improving the clustering accuracy via correcting the broken links of the representation graph.
\end{abstract}


\section{Introduction}
Clustering is one of the challenging problems in the area of machine learning and data analysis~\cite{xu2005survey}, for which 
unsupervised methods try to discover the hidden structure of the data. 
On the other hand, sparse coding algorithms aim for finding a latent representation of data points based on a weighted combination of sparsely selected base vectors \cite{rubinstein2008efficient}.
Such a sparse representation has the potential to capture the essential characteristics of the data including its hidden structure \cite{kim2010sparse}.
Therefore, in recent years, several studies have tried and succeeded in using sparse coding models for clustering purposes~\cite{liu2013robust,you2016scalable,xu2016novel}. 
Calling the weighting coefficients sparse codes, the clustering phase is applied to the learned sparse codes using common clustering methods such as spectral clustering~\cite{yang2014data}.

An important group of sparse coding methods for clustering is called sparse subspace clustering algorithms (SSC) \cite{elhamifar2013sparse}. 
Assuming the data is distributed on a union of linear subspaces,
SSC methods focus on obtaining self-expressive representations, such that each data point could be represented by using other samples from its cluster (subspace) ~\cite{Cheng:2010:LLG:1820776.1820778,liu2013robust}.
There are considerable variations in the structure of existing SSC algorithms \cite{vidal2014low,Gao2012,you2016scalable}, which leads to different optimization schemes.

From another point of view, some of the sparse coding approaches restrict the sparse codes to non-negative values to obtain a more interpretable representation for the data, especially when the data is related to biological models~\cite{hoyer2003modeling}. Such non-negativity also often results in a better construction of the subsequent clustering graph~\cite{xiao2016robust,zhuang2012non}.

Benefiting from kernel functions, it could be possible to transfer data to a high-dimensional space in which clusters are more separable. 
Hence,
a subset of SSC algorithms focused on developing kernel-based SSC methods \cite{patel2014kernel,bian2016kernelized,xiao2016robust} which typically achieve higher clustering accuracies in comparison to their vectorial versions.
\subsubsection*{Contributions:}
In this work, we propose a non-negative SSC algorithm with a unique structure. The method combines nuclear-norm with a local-separability objective term. In addition, it preserves the affine representation of data in the latent space in accordance with an affine assumption about the underlying subspaces.
Accordingly, our explicit contributions are as follows:
\begin{itemize}
	\item We introduce and add a novel objective term to the problem which focuses on increasing local separability of data. This term is used in an unsupervised way, and it affects the sparse representation of data to have a better cluster separability.
	\item An efficient post-processing method is introduced regarding the negative effect of sparse coding redundancies on clustering performance.
	\item Our algorithm is also extended to the nonlinear version via incorporating a kernel function in its framework.
\end{itemize}

In the next section, we briefly review SSC algorithms.
Afterward, we present our proposed approaches in Sec. \ref{sec:method} and the optimization procedure in Sec. \ref{sec:optim}. Then, we carry out empirical evaluations in Sec.~\ref{secexp}, and make the conclusion in the final section. 
\section{Related Works}
Consider the data matrix $\nY=[\ny_1,...,\ny_N] \in \mathbb{R}^{d\times N}$ which lies in the union of $n$ linear subspaces $\cup_{l=1}^n \MC{S}_l$ each with the dimension of $\{d_l\}_{l=1}^n$. Subspace clustering tries to cluster the data such that each cluster $i$ contains samples lying in one individual subspace $\MC{S}_i$. Therefore, each data point $\ny_i$ can be represented by other data points in $\nY$ as a linear combination $\ny_i \approx \nY\nx_i$. Focusing on the sparseness of the coding vectors $\nx_i$, subspace sparse clustering \cite{elhamifar2013sparse} can be formulated as 
\begin{equation}
\begin{array}{ll}
\underset{\nX} {\min} \|\nX\|_0\quad s.t. ~~\nY=\nY\nX, \nxs_{ii}=0~,~\forall i
\label{eq:ssc}
\end{array}
\end{equation}
where $\nX$ is the matrix of sparse codes, $\nxs_{ii}$ points to diagonal elements of $\nX$, and $\|.\|_0$ denotes the cardinality norm.  
It is assumed each resulting $\nx_i$ from Eq. \ref{eq:ssc} represents $\ny_i$ using only data points from the subspace in which $\ny_i$ lies as well. 
In that case, computing an affinity matrix $\MB{A}=\left|\nX\right|^{\top}+\left|\nX\right|$ which represents
the pairwise similarities of data points, and using it in graph-based methods such as spectral clustering should identify the clusters.
However, the problem in Eq.\ref{eq:ssc} is NP-hard to solve \cite{elhamifar2013sparse} in its original format. As a solution, $\|.\|_0$ can be relaxed into other norms. For instance \cite{elhamifar2013sparse,patel2014kernel,bian2016kernelized,Gao2012} use the $l1$-norm to achieve sparse $\nX$, while \cite{you2016scalable} aims for the approximate solution of Eq.~\ref{eq:ssc} while having $\|\nx_i\|_0 \leq \nT$. 
Another group of SSC methods \cite{vidal2014low,xiao2016robust,liu2013robust,zhuang2012non} focuses on shrinking the nuclear norm $\|\nX\|_*$ and making $\nX$ low-rank to better represent the global structure of data. 
Among SSC algorithms, \cite{elhamifar2013sparse,patel2014kernel} enforced $\nX$ to provide affine representations by using the constraint $\nX^\top \vec{\MB{1}}=\vec{\MB{1}}$ based on the idea of having the data points lying on an affine combination of subspaces.
Despite continuous improvements in clustering results of aforementioned SSC methods, 
there is no direct link between the quality of the sparse coding part and the subsequent clustering goal. Consequently, they suffer from performance variations across different datasets and high sensitivity of their results to the choice of parameters.

On the other hand, another group of algorithms called Laplacian sparse coding encourage the sparse coefficient vectors $\nx_i$ related to each cluster to be as similar as possible~\cite{Gao2012,yang2014data}. 
In their SSC formulation (Eq. \ref{optlap}) they employ a similarity matrix $\MB{W}$ in which each $w_{ij}$ measures the pair-wise similarity between a pair $(\ny_i,\ny_j)$.
\begin{equation}
\begin{array}{ll}
\underset{\nX}{\min}&\|\nY-\nY\nX\|_F^2+\lambda \|\nX\|_1+\frac{1}{2}\sum_{i,j} w_{ij}\|\nx_i-\nx_j\|_2^2 
\qquad s.t. ~\nxs_{ii}=0 ~,~ \forall i
\label{optlap}
\end{array}
\end{equation}
Nevertheless, the optimization frameworks like this suffer from two issues:
\begin{enumerate}
	\item Columns of $\nX$ are forced to become similar to each other while the similarity matrix is used as the weighting coefficients. 
	Hence, at best the sparse codes $\nx_i$ obtain a distribution similar to the neighborhoods in $\MB{W}$. Consequently, their performance is comparable to kernel-based clustering with direct use of the kernel information.
	\item Although Eq. \ref{optlap} tries to decrease the intra-cluster distances, the inter-cluster structure of data is ignored in such frameworks; however, typically both of these terms have to be adopted when focusing on the separability of clusters. 
\end{enumerate}
Contrary to the previous works, our algorithm benefits from a clustering-based objective term in its framework. Therefore, its resulting sparse codes are more suitable for the clustering purpose. 
In addition, our post-processing technique can contribute to non-negative SSC methods such as \cite{li2017graph,xiao2016robust,zhuang2012non} to improve their latent representations.  
\section{Proposed Non-Negative SSC algorithm}\label{sec:method}
In this section, we introduce our proposed SSC algorithms NLSSC and NKLSSC. Although they are explained in individual subsections, NKLSSC is the kernel extension of NLSSC which is optimized similarly to NLSSC's optimization. 
\subsection{Non-Negative Local Subspace Sparse Clustering}
We formulate our non-negative local SSC algorithm (NLSSC) using the following self-representative framework:
\begin{equation}
\begin{array}{ll}
\underset{\nX}{\min}&\|\nX\|_*+\frac{\lambda }{2} \| \nY-\nY\nX\|_F^2 +\mu\mathcal{E}_{lsp}(\nX,\nY)\\
\mathrm{s.t} & 
~ \nX^{\top}\vec{\MB{1}}=\vec{\MB{1}},
\nxs_{ij} \ge 0  , ~~\nxs_{ii}=0~,~\forall ij
\label{eq:nnlssc}
\end{array}
\end{equation}
where $\nxs_{ii}=0$ prevents data points from being represented by own contributions.
The constraint $\nX^{\top}\vec{\MB{1}}=\vec{\MB{1}}$ focuses on the affine reconstruction of data points which coincides with having the data lying in an affine union of subspaces $\MC{S}_l$. 
The nuclear norm regularization term $\|\nX\|_{*}=trace(\sqrt{\nX^*~\nX})$ is employed to ensure the sparse coding representations are low-rank. This helps the sparse model to better capture the global structure of data distribution.
The non-negativity constraint on $\nxs_{ij}$ is employed to enforce the data combinations to happen mostly between similar samples.
The novel term $\mathcal{E}_{lsp}(\nX,\nY)$ is a loss function which focuses on the local separability of data points in the coding space based on values of $\nX$.
Accordingly, scalars $\lambda$ and $\mu$ are constants which control the contribution of the objective terms.
The goal of having $\mathcal{E}_{lsp}(\nX,\nY)$ in the SSC model is to reduce intra-cluster distance and increase inter-cluster distance. To do so in an unsupervised way, we define
\begin{equation}
\mathcal{E}_{lsp}(\nX,\nY):=\frac{1}{2}\sum_{i,j} \big[ w_{ij}\|\nx_i-\nx_j\|_2^2 
+ b_{ij} ({\nx_i}^\top \nx_j)\big]
\label{eq:olsp}
\end{equation}
in which the binary regularization weighting matrices $\MB{W}$ and $\MB{B}$ are computed as
\begin{equation}
w_{ij}= 
\begin{cases}
1, & \text{if } \ny_j \in \MC{N}_i^k\\
0,              & \text{otherwise}
\end{cases}
,\qquad
b_{ij}= 
\begin{cases}
1, & \text{if } \ny_j \in \MC{F}_i^{k}\\
0,              & \text{otherwise}
\end{cases}
\label{eq:weight}
\end{equation}
The two sets $\MC{N}_i^k$ and $\MC{F}_i^k$ refer to the $k$-nearest and $k$-farthest data points to $\ny_i$, and are determined via computing Euclidean distance $\|\ny_i-\ny_j\|_2$ between each $\ny_i$ and $\ny_j$. 
Defining $\MC{D}(\MB{W},\nX):=\sum_{i,j} w_{ij}\|\nx_i-\nx_j\|_2^2$ 
and $\MC{H}(\MB{B},\nX):=\sum_{i,j} b_{ij} ({\nx_i}^\top \nx_j)$, 
the first part reduces the distance between $(\nx_i,\nx_j)$ if they belong to $\mathcal{N}_i^k$ while the latter focuses on incoherency of each pair of $(\nx_i,\nx_j)$ if they are members of $\mathcal{F}_i^k$.
The following explains the effect of $\mathcal{E}_{lsp}$ on the separability of the clusters in the coding space.

Assuming there exist the labeling scalars $\{l_i\}_{i=1}^N \in \MBB{R}$,
we prefer $\ny_i$ and members of $\MC{N}_i^k$ to belong to the same class while the set $\MC{F}_i^k$ to contain data from other clusters.
We define $\MB{W}_c$ and $\MB{W}_m$ such that $\MB{W}=\MB{W}_c+\MB{W}_m$, and 
they respectively denote the correct and wring assignments regarding the label information $\nh$. 
more precisely, if $w(i,j)=1$ then in case $l_i=l_j$ we have ${w_c}(i,j)=1$, 
otherwise ${w_m}(i,j)=1$. The rest of the entries in $({\MB{W}_c},{\MB{W}_m})$ are set to zero.
\begin{definition} 
	The neighborhoods in $\nY$ are cluster representative to the order of 
	$o_r$, if $\exists k\in \mathbb{N}:\|{\MB{W}}_c\|_0/\|\MB{W}_m\|_0=o_r,~\text{and}~ o_r<1$.
	\label{def:cr}
\end{definition}
Definition \ref{def:cr} means that in the neighborhoods of data samples in $\nY$ there are more points of the same class than of different ones. 
\begin{proposition}
	Minimizing $\mathcal{E}_{lsp}$ in Eq. (\ref{eq:olsp}) makes columns of $\nX$ to be better locally separable regarding the underlying classes, 
	if the neighborhoods in $\nY$ are cluster representative with a sufficiently small $o_r$. 			
	\begin{proof}\{\textit{sketch}\}
		Eq.~\ref{eq:olsp} can be rewritten as 
		$$\mathcal{E}_{lsp}=\DW{W}{c}{}+\DW{W}{m}{}+\MC{H}(\MB{B},\nX)$$		
		Therefore,  
		$\nX^* = \underset{\nX}{\argmin}~\MC{E}_{lsp}$		
		generally works in favor of decreasing $\DW{W}{c}{}$ and $\MC{H}(\MB{B},\nX)$ compared to an initial $\nX^0$.
		\\Consequently, a small $\DW{W}{c}{}$ leads to compact same-label neighborhoods in $\nX^*$, 		  
		and decreasing $\MC{H}(\MB{B},\nX)$ generally increases $\DW{B}{}{}$ and 
		more provides a more localized structure for $\nX^*$.  				
		\\Denoting $\Delta\DW{W}{}{*}:=\DW{W}{}{*}-\DW{W}{}{0}$,
		according to the definition~\ref{def:cr}, 
		${\Delta\DW{W}{m}{*}}/{\Delta\DW{W}{c}{*}}$ 
		is a decreasing function of $1/o_r$.			
		\\Hence, the smaller $o_r$ becomes the more columns of $\nX^*$ can be locally separated from data samples of the other classes ($\MB{W}_m$) in their neighborhoods.			
	\end{proof}	
	\label{prop1}	
\end{proposition}

Proposition \ref{prop1} shows 
the effect of minimizing the loss term $\mathcal{E}_{lsp}$ on having localized and condense neighborhoods in the sparse codes $\nX$ by making the sparse codes of the neighboring samples more similar (identical in ideal case) while making those of far away points incoherent (orthogonal in ideal case).   
It also provides the desired condition by which the local neighborhoods in $\nX$ 
can better respect the class labels $\vec{l}$ and leading to a better alignment between $\nX$ and the underlying 
subspaces. 
%
\textbf{Note:} Here we referred to $\nh$ only to explain the reason behind our specific model design; however, 
the algorithm does need the labeling information in any of its steps.
\subsection{Clustering based on $\nX$}
Similar to other SSC algorithms, the resulted sparse coefficients
are used to construct an adjacency matrix $\MB{A}=\nX+\nX^{\top}$ defining a a sparse representation graph $\MC{G}$. This undirected graph consists of weighted connections between pairs of $(\ny_i,\ny_j)$. 
Therefore, $\MB{A}$ is used as the affinity matrix in the spectral clustering algorithm \cite{yang2014data} to find the data clusters.
\subsection{Link-Restore} 
After constructing the affinity matrix based on the resulting $\nX$, 
it is desired to have positive weights in the representation graph $\MC{G}$ between every two points of a cluster. However, in practice, it is possible to see non-connected nodes (broken links) even inside condense clusters. This happens due to the redundancy issue related to sparse coding algorithms.
In Eq. \ref{eq:nnlssc}, 
$\nY$ is used as an over-complete dictionary for reconstruction of each $\ny_i$, 
therefore we can assume $\ny_i\approx \nY\nx_i$.
Nevertheless, as a common observation in sparse coding models the solution for the value of $\nx_i$ is suboptimal because of the utilized $\|\nx_i\|_p$ relaxations. Thus for $\ny_s$ as a close data point to $\ny_i$, it is possible to have 
$\ny_s\approx \nY \nx_s$, but with a big ${\nx}_i^\top{\nx}_s$. 
This means $\nx_i$ and $\nx_s$ are not similar in the entries. 
Consequently, 
$a_{ij}$ can be small resulting from dissimilar $\nx_i$ and $\nx_s$, albeit $\ny_i$ and $\ny_s$ are very similar.
%
\begin{algorithm}[!b]
	\small
	\SetAlgoLined
	\KwInput{Sparse code $\nx$, data matrix $\nY$, threshold $\tau \in [0,1]$}
	\KwOutput{Corrected $\nx$ by restoring its connections to other data points} 
	\Kwinit{$I=\{i\mid \nxs_i\neq 0\}$ \scriptsize (except index of $\ny$)} 
	\KwProc{$\{$over all elements $i \in I$ $\}$}
	\quad $\hat{\nx}=\nx$,
	\quad $\bar{I}:=\{s\mid (\ny_s^\top\ny_s-2\ny_i^\top\ny_s)<(\tau -1)\ny_i^\top\ny_i ~,~ \nxs_s=0   \}$\\ \label{line:is}
	\quad $\hat{\nxs}_i={\nxs}_i (\ny_i^\top\ny_i/{\sum_{s\in \{\bar I \cup i\}}{\ny_i^\top\ny_s}})$\\ \label{line:yi}
	\quad $\hat{\nxs}_s=\hat{\nxs}_i(\ny_i^\top\ny_s/{\ny_i^\top\ny_i})~,~\forall s\in \bar{I}$\\ \label{line:ys}
	\quad $\nx=\hat{\nx}$,\quad $I=I\backslash\{i\}$\\
	\caption{Link-Restore post-processing}
	\label{alg:graph}
\end{algorithm}
As a workaround to the mentioned issue, we propose the {Link-Restore} method (Algorithm~\ref{alg:graph}) as an effective step regarding these situations. It acts as a post-processing step on the obtained $\nX$ before application of spectral clustering.
Link-restore corrects entries of each $\nx$ by restoring the broken connections between $\ny$ and other points in the dataset. To do so, 
it first obtains the current set of data points connected to $\ny$ as $I=\{i\mid \nxs_i\neq 0\}$,
where $\nxs_i$ denotes $i$-th entry in vector $\nx$.
Then for each $\nx_i$ that $i \in I$, the algorithm 
collects the indices $\bar{I}$ of data points which are close to $\ny_i$ but not used in the sparse code of $\ny$  (line~\ref{line:is}).
To that aim, for each $\ny_s \in \bar{I}$ the criterion ${\|\ny_i-\ny_s\|^2_2}/{\|\ny_i\|^2_2}<\tau$ should be fulfilled,
where $0\leq \tau \leq 1$.
Then in order to incorporate members of $\bar{I}$ into $\nx$, the entry $\nxs_i$ is projected to $\bar{I}\cup i$ based on the value of  ${\ny_i^\top\ny_s}/{\ny_i^\top\ny_i} ~~\forall s\in \bar{I}$
while also maintaining the affinity constraint on $\nx$ (lines~\ref{line:yi}-\ref{line:ys}).
It is important to point out that the pre-assumption for the above is that $\nxs_{i} \ge 0~~\forall i$. Therefore link-restore method can be assumed as a proper post-processing method for \textit{non-negative} subspace clustering algorithms.
\subsection{Kernel Extension of NLSSC}
Assume $\Phi:\MBB{R}^d\rightarrow\MBB{R}^m$ is an implicit nonlinear mapping to a Reproducing Kernel Hilbert Space (RKHS) such that $m \gg d$.
Thus, there exists a kernel function
$\K(\ny_i,\ny_j)=\Phi(\ny_i)^{\top}\Phi(\ny_j)$. 
Doing so, we can benefit from the non-linear characteristics of this implicit mapping to obtain better representation for the data.
Accordingly, we can reformulate our NLSSC method (Eq. \ref{eq:nnlssc}) into its kernel extension as the non-negative local kernel SSC algorithm (NLKSSC):
\begin{equation}
\begin{array}{ll}
\underset{\nX}{\min}&\|\nX\|_*+\frac{\lambda }{2} \| \Phi(\nY)-\Phi(\nY)\nX\|_F^2 +\mu\mathcal{E}_{lsp}(\nX,\Phi(\nY))\\
\mathrm{s.t} & ~ \nX^{\top}\vec{\MB{1}}=\vec{\MB{1}}, ~\nxs_{ij} \ge 0  , ~~\nxs_{ii}=0~,~\forall ij
\label{eq:nklssc}
\end{array}
\end{equation}
Comparing to Eq. \ref{eq:nnlssc}, the second term in the objective of Eq.\ref{eq:nklssc} means a self-representation in the feature space, and the local-separability term ($\MC{E}_{lsp}$) is equivalent to the one used in \ref{eq:nnlssc}. However, $\MB{W}$ and $\MB{W}_m$ in $\MC{E}_{lsp}$ are computed based on the entries $\K(\ny_i,\ny_j)$ which 
directly indicate the pair-wise similarity of each data $\ny_i$ to its surrounding neighborhood.
The benefit of having a kernel representation of $\nY$ is that a proper kernel function facilitates the validity of the pre-assumption for Proposition \ref{prop1}, which leads to the more efficient role of $\MC{E}_{lsp}$.  
As we see in Sec. \ref{sec:optim}, we can use the same optimization regime for both NLSSC and NLKSSC.
In addition, the lines~\ref{line:is}-\ref{line:ys} of the link-restore algorithm can be implemented using the above dot-product rule.
\section{Optimization Scheme of Proposed Methods}\label{sec:optim}
Putting Eq. \ref{eq:olsp} into Eq. \ref{eq:nnlssc} the following optimization framework is derived
\begin{equation}
\begin{array}{ll}
\underset{\nX}{\min}&\|\nX\|_*+\frac{\lambda }{2} \| \nY-\nY\nX\|_F^2 
+\frac{\mu}{2}\sum_{i,j} \big[ w_{ij}\|\nx_i-\nx_j\|_2^2 
+ b_{ij} ({\nx_i}^\top\nx_j)\big]\\
\mathrm{s.t} & ~ \nX^{\top}\vec{\MB{1}}=\vec{\MB{1}}, ~\nxs_{ij} \ge 0  , ~~\nxs_{ii}=0~,~\forall ij
\label{eq:optim}
\end{array}
\end{equation}
To simplify the 3rd loss term in (\ref{eq:optim}), we symmetrize $\MB{W}\rightarrow \frac{\MB{W}+\MB{W}^{\top}}{2}$ and do the same for $\MB{B}$.
Then according to \cite{von2007tutorial} we compute the Laplacian matrix $\MB{L}=\MB{D}-\MB{W}$, where $\MB{D}$ is a diagonal matrix such that $d_{ii}=\sum_{j}w_{ij}$. 
Then, with simple algebric operations we can rewrite $\mathcal{E}_{lsp}(\nX,\nY)=\Tr(\nX \MB{L} \nX^\top)
+\frac{1}{2}\Tr(\nX \MB{B} \nX^\top)$,
and reformulate Eq.~\ref{eq:optim} as: 
\begin{equation}
\begin{array}{ll}
\underset{\nX}{\min}&\|\nX\|_*+\frac{\lambda }{2} \| \nY-\nY\nX\|_F^2 +\mu \Tr(\nX\hat{\MB{L}}\nX^{\top}) \\
\mathrm{s.t} & ~ \nX^{\top}\vec{\MB{1}}=\vec{\MB{1}}, ~\nxs_{ij} \ge 0  , ~~\nxs_{ii}=0~,~\forall ij
\label{opttr}
\end{array}
\end{equation}
where $\Tr(.)$ is the trace operator and 
$\hat{\MB{L}}=(\MB{L}+\frac{1}{2} \MB{B})$. 
The objective of Eq. \ref{opttr} is sum of convex functions (trace, inner-product, and convex norms), therefore the 
optimization problem is a constrained convex problem
and can be solved using the alternating direction method of multipliers (ADMM) \cite{boyd2011distributed} as presented in Algorithm \ref{alg:admm}.
Optimizing Eq.~\ref{opttr} coincides with minimizing the following augmented Lagrangian which is derived by adding its constraints as penalty terms in the objective function.
\begin{equation}
\begin{array}{ll}
\MC{L}_{\rho}&(\nX,\nX_+,\MB{U},\alpha_+,\alpha_U,\vec{\alpha_\MB{1}})
=\|\MB{U}\|_*+\lambda \MC{E}_{rep}(\nY,\nX)+\mu\MC{E}_{lsp}(\nY,\nX)\\
&+\frac{\rho}{2}\|\nX-\nX_+ \|_F^2+\Tr(\alpha_{+}^\top(\nX-\nX_+))
+\frac{\rho}{2}\|\nX-\MB{U} \|_F^2\\
&+\Tr(\alpha_{U}^\top(\nX-\MB{U}))
+\frac{\rho}{2}\|\nX^\top\vec{\MB{1}}-\vec{\MB{1}} \|_2^2
+\langle\vec{\alpha_\MB{1}},\nX^\top\vec{\MB{1}}-\vec{\MB{1}}\rangle
\end{array}
\label{eq:lagran}
\end{equation}
in which $\MC{E}_{rep}:=\frac{1}{2} \| \nY-\nY\nX\|_F^2$, 
and matrices $(\nX_+,\MB{U})$ are axillary matrices related the non-negativity constraint and the term $\|\nX\|_*$.
Eq~\ref{eq:lagran} contains the Lagrangian multipliers $\alpha_+,\alpha_U \in \MBB{R}^{N\times N}$ and $\vec{\alpha_\MB{1}} \in \MBB{R}^{N}$,
and the penalty parameter $\rho \in \MBB{R}^+$.
Minimizing $\MC{L}_{\rho}$ Eq.\ref{eq:lagran} 
is carried out in an alternating optimization framework, such that at each step of the optimization all of the parameters $\{\nX,\nX_+,\MB{U},\alpha_+,\alpha_U,\vec{\alpha_\MB{1}}\}$ are fixed except one.
Therefore, the updating steps are described as follows.
\newline
\textit{\textbf{Updating $\nX$}}: 
At iteration $t$ of ADMM, via fixing $\nX_+^t,\MB{U}^t,\alpha_+^t,\alpha_U^t,\vec{\alpha}_\MB{1}^t$, the matrix $\nX^{t+1}$ is updated as the solution to this Sylvester linear system of equations \cite{kirrinnis2001fast}
\begin{equation}
[2\lambda\nY^\top\nY+2\rho \MB{I}+\vec{\MB{1}}\vec{\MB{1}}^\top]\nX^{t+1}+\nX^{t+1}[2\mu\hat{\MB{L}}]
=\rho[\nX_+^t+\MB{U}^t+\vec{\MB{1}}\vec{\MB{1}}^\top]-\alpha_{\MB{U}}^t-\alpha_{+}^t
-\vec{\MB{1}}{\vec{\alpha^t}_{\MB{1}}}^\top
\label{eq:up_x}
\end{equation}
\textit{\textbf{Updating $\MB{U}$}}:
Updating $\MB{U}^{t+1}$ which is associated with $\|\nX\|_*$ can be done
via fixing other parameters and using the singular value thresholding method \cite{cai2010singular} as  
$\MB{U}^{t+1}=\MC{T}_{1/\rho}(\nX)$
where term $\MC{T}(.)$ is the thresholding operator from \cite{cai2010singular}(Eq. 2.2).
\newline
\begin{algorithm} [!b]
	\caption{Optimization Scheme of NLSSC} 
	\label{alg:admm} 
	\LinesNumberedHidden	
	\KwInput{$\nY,\lambda,\mu,k,\Delta_\rho=0.1,\rho_{max}=10^6$}
	\KwOutput{Sparse coefficient matrix $\nX$}
	\Kwinit{Compute $\{\MB{W},\MB{B},\hat{\MB{L}}\}$. Set all $\{\nX_+,\nX,\MB{U},\alpha_+,\alpha_U,\vec{\alpha_\MB{1}}\}$ to zero}
	\Repeat{Convergence Criteria is fulfilled}
	{
		Updating $\nX$ by solving Eq.~\ref{eq:up_x}\\
		Updating $\MB{U}$ based on \cite{cai2010singular}(Eq. 2.2)\\
		Updating $\nX_+,\alpha_+,\alpha_U,\vec{\alpha_\MB{1}}$ based on Eq.~\ref{eq:up_rest}
	}
\end{algorithm}	
\textit{\textbf{Updating $\nX_+,\alpha_+,\alpha_U,\vec{\alpha_\MB{1}},\rho$}}:
The matrix $\nX_+$ and the multipliers are updated using the following projected gradient descent and 
gradient ascent steps respectively
\begin{equation}
\begin{array}{ll}
\nX_+^{t+1}=\max(\nX+\frac{1}{\rho} \alpha_+,0),&
\qquad\alpha_+^{t+1}=\alpha_+^{t}+\rho(\nX-\nX_+)\\
\vec{\alpha}_{\MB{1}}^{t+1}=\vec{\alpha}_{\MB{1}}^{t}+\rho(\nX^\top\vec{\MB{1}}-\vec{\MB{1}}),&
\qquad\rho^{t+1}=\min(\rho^t(1+\Delta_\rho),\rho_{max})
\end{array}
\label{eq:up_rest}
\end{equation}
in which $(\Delta_\rho,\rho_{max})$ are the update step and higher bound of $\rho$ respectively.
\newline
\textit{\textbf{Convergence Criteria}}:
The algorithm reaches its convergence point when for a fixed $\epsilon>0$, 
$\|\nX^t-\nX^{t-1}\|_\infty \leq \epsilon$, 
$\|\nX_+^t-\nX^t\|_\infty \leq \epsilon$, 
$\|\MB{U}^t-\nX^t\|_\infty \leq \epsilon$, 
and $\|{\nX^t}^\top \vec{\MB{1}}-\vec{\MB{1}}\|_\infty \leq \epsilon$.
\subsubsection*{Optimizing NLKSSC:}
The kernel-based algorithm (NLKSSC) is optimized also using Algorithm \ref{alg:admm} while the kernel trick 
$\Phi(\ny_i)^\top\Phi(\ny_j)=\K(\ny_i,\ny_j)$ 
is applied to replace $\nY^\top\nY$ by $\K(\nY,\nY)$ in Eq. \ref{eq:up_x}, and to kernelize the link-restore algorithm as well.
\section{Experiments}\label{secexp}
For empirical evaluation of our proposed NNLSSC and NLKSSC algorithms, we implement them on 4 different widely-used benchmarks of clustering datasets:
\begin{itemize}
	\item Hopkins155 \cite{tron2007benchmark}: Segmentation of 156 video sequences with a setup similar to \cite{elhamifar2013sparse}. 
	\item COIL-20 \cite{nene1996columbia}: A dataset of 1440 gray-scale images of 20 different objects with the pixel size of $32\times 32$.
	\item Extended YaleB\cite{georghiades2001few}:
	Contains frontal face images taken from 38 subjects with the average of 64 samples per subject. Feature extraction is done based on \cite{vidal2014low}.
	\item AR-Face \cite{martinez1998ar}: An image dataset including more than 4000 frontal faces of 126 different subjects. We use 2600 images from 100 subjects and use the pre-processing procedure from \cite{xiao2016robust}.
\end{itemize}
The basis of evaluation is the clustering error as $CE=\frac{\text{\# of miss-clustered samples}}{\text{\# of data samples}}$ using the posterior labeling of the clusters
 and the normalized mutual information ($NMI$) \cite{ana2003robust}. 
For each method, an average $CE$ is calculated over 10 runs of the algorithm.
$NMI$ measures the amount of information shared between the clustering result and the ground-truth which lays in range of $\big[0,1\big]$ with the ideal score of 1.
Based on the common practice in the literature, we use average $CE$ along with its median value for the Hopkin155 dataset.

We compare our algorithms' performance to baseline methods
{SSC}~\cite{elhamifar2013sparse},
{LRSC} \cite{vidal2014low},
{SSOMP}~\cite{you2016scalable},
{S$^3$C} \cite{li2015structured},
{GNLMF} \cite{li2017graph},
{KSSC}~\cite{patel2014kernel}
{KSSR} \cite{bian2016kernelized} and
{RKNNLRS} \cite{xiao2016robust}.
These algorithms are selected from major sparse coding-based clustering approaches, among which 
{KSSC}, {KSSR}, and {RKNNLRS} are kernel-based methods. The spectral clustering step of the baselines is performed via using the correct number of clusters.

To compute the kernels required for kernel-based we use Histogram Intersection Kernel (HIK) as in \cite{wu2009beyond} for AR dataset as it is a proper choice regarding its frequency-based features \cite{xiao2016robust}. 
For the implementations on the rest of the datasets we adopted the Gaussian kernel $\K (x,y)=exp(-\frac{\| x-y \|^2}{\sigma})$, where $\delta$ is the average of $\| \ny_i-\ny_j \|^2$ over all data samples.
\subsection{Parameter Settings}
In order to tune the parameters $\lambda,\mu,k$ we utilize a grid-search method. We do the search for $\lambda$ in the range of $\{1,1.5,...,7\}$, for $\mu$ in the range of $\{0.1,0.2,...,1\}$ and $k$ in $\{3,4,...,8\}$.
We implement a similar parameter search for the baselines to find their best settings.
Although for the link-restore parameter, $\tau=0.2$ generally works well, one can do a separate grid-search for $\tau$. 
\begin{table*}[!b]
	\caption{Average clustering error ($CE$) and $NMI$ for YALE, COIL20, AR datasets. $CE$ and its median value for Hopkins155-(2 motions and 3 motions) datasets.}
	\vspace{-0.5cm}
	\label{tab:result}	
	\LARGE
	\begin{center}
	\resizebox{1\textwidth}{!}{%
		\begin{tabular}{|l|c|c||c|c||c|c||c|c||c|c|} 
			\hline
			\multirow{2}{*}{Methods} & \multicolumn{2}{c||}{YALE B} & \multicolumn{2}{c||}{COIL20}  & \multicolumn{2}{c||}{AR-Face}& 
			\multicolumn{2}{c||}{Hopkins-2m} & \multicolumn{2}{c|}{Hopkins-3m}\\
			\cline{2-11}
			& $CE$ & $NMI$ & $CE$ & $NMI$ & $CE$ & $NMI$ & $CE$ & med. & $CE$ & med.\\
			\hline
			{SSC}~\cite{elhamifar2013sparse} &0.1734&0.8902&0.1737&0.9104&0.1065&0.9103& 0.0289 &0 &0.0663 &0.0114 \\
			\hline			
			{LRSC}\cite{vidal2014low}        &0.3136 &0.7340&0.2943&0.7838&0.0938&0.9037  & 0.0369&0.2127 &0.0746 &0.0245 \\
			\hline			
			{SSOMP}\cite{you2016scalable}    & 0.3214&0.6792&0.7652&0.5274&0.1012&0.8353 & 0.1432 & 0.0328& 0.1973& 0.1504 \\
			\hline			
			{S$^3$C}\cite{li2015structured}  &0.1565 &0.9104&0.1635&0.9063&0.0897&0.9117 & 0.0263& 0& 0.0527& 0.0089\\
			\hline			
			{GNLMF}\cite{li2017graph}        & 0.3074&0.4172&0.3972&0.6421&0.1544&0.8769 &0.1052 & 0.0216& 0.1239& 0.0841 \\
			\hline			
			{KSSC}\cite{patel2014kernel}     & 0.1504&0.8907&0.1833&0.9039&0.0678&0.9241 & 0.0275& 0& 0.0584& 0.0096\\
			\hline			
			{KSSR}\cite{bian2016kernelized}  &0.1598&0.8864&0.1983&0.9027&0.0742&0.8983 & 0.0437& 0.6121& 0.0756& 0.0151\\
			\hline			
			{RKNNLRS}\cite{xiao2016robust}   &0.1493&0.9035&0.1672&0.9126&0.0886&0.9131 & 0.0254& 0& 0.0512& 0.0087\\
			\hline			
			\hline
			\textbf{NLSSC}\large(Proposed)       &0.1242&0.9146&\textbf{0.1409}&\textbf{0.9254}&0.0832&0.9125& 0.0189& 0& 0.0427 & 0.0079\\
			\hline			
			\textbf{NLKSSC}\large(Proposed)      &\textbf{0.1107}&\textbf{0.9163}&0.1528&0.9147&\textbf{0.0542}&\textbf{0.9364} & \textbf{0.0122}& 0& \textbf{0.0331}& \textbf{0.0065}\\
			\hline
		\end{tabular}
	}
	\footnotesize
	The best result (\textbf{bold}) is according to a two-valued t-test at a $5\%$ significance level.		
	\end{center}
\end{table*}
\subsection{Clustering Results}
According to the results summarized in Tables. (\ref{tab:result}), the proposed methods outperformed the benchmarks regarding the clustering error. Comparing NLKSSC to NLSSC, the kernel-based algorithm resulted in a smaller $CE$ compared to NLSSC (except for COIL20), which shows that the kernel-based framework was able to better represent cluster distributions.
Regarding the COIL20 dataset, via comparing kernel-based methods to other baselines, it can be concluded that the utilized kernel function was not strongly effective for cluster-based representation of the dataset. However, NLKSSC still outperformed other baselines due to the effectiveness of its sparse subspace model.

Among other methods, S$^3$C, RKNNLRS, and KSSC have comparable results, especially for the Hopkins dataset. 
This means, although KSSC and RKNNLRS benefited from kernel representation, the S$^3$C algorithm was relatively effective regarding capturing the data structure. 
However, KSSR presented low performance even in comparison to vectorial methods such as SSC and LRSC. 
This behavior is due to lack of having any strong regularization term in its model regarding the subspace structure of data.  
Among non-negative methods, GNLMF performance is relatively below average. This may suggest that its NMF-based structure is not suitable for grasping cluster distribution in comparison to self-representative methods. 
On the other hand, RKNNLRS performance shows that its non-negative model is more effective for clustering purposes compared to NMF-based models. 
Comparing NLSSC (the proposed algorithm) to other baselines with low-rank regularizations in their models,
we can conclude that proper combination of the locality term and the affine constraints aided NLSSC to obtain higher performance. 
The same conclusion can be derived via comparing NLSSC/NLKSSC to KSSC as an affine subspace clustering algorithm.
\begin{table*}[!b]
	\caption{Application of the link-restore method on the non-negative approaches.}
	\label{tab:link}
	\vspace{-0.5cm}	
	\LARGE
	\begin{center}
		\resizebox{1\textwidth}{!}{%
			\begin{tabular}{|l|c|c||c|c||c|c||c|c||c|c|} 
				\hline
				\multirow{2}{*}{Methods} & \multicolumn{2}{c||}{YALE} & 
				\multicolumn{2}{c||}{COIL20}  & \multicolumn{2}{c||}{AR}& 
				\multicolumn{2}{c||}{Hopkins-2m} & \multicolumn{2}{c|}{Hopkins-3m}\\
				\cline{2-11}
				& $CE$ & $NMI$ & $CE$ & $NMI$ & $CE$ & $NMI$ & $CE$ & median & $CE$ & median\\
				\hline
				{GNLMF-link}\cite{li2017graph}        &0.2514& 0.6564&0.2674&0.7161&0.1251&0.8846&0.0793&0.0147&0.1025&0.0649\\
				\hline			
				{RKNNLRS-link}\cite{xiao2016robust}   &0.1237&0.9103&0.1602&0.9137&0.0823&0.9135&0.0230&{0}&0.0469&0.0081\\
				\hline			
				\hline
				\textbf{NLSSC-link}       &0.1027&0.9182&\textbf{0.1409}&\textbf{0.9254}&0.0776&0.9153&0.0189&{0}&0.0392&0.0064\\
				\hline			
				\textbf{NLKSSC-link}      &\textbf{0.0842}&\textbf{0.9326}&0.1523&0.9148&\textbf{0.0482}&\textbf{0.9381}&\textbf{0.0122}&{0}&\textbf{0.0301}&\textbf{0.0054}\\
				\hline
			\end{tabular}
		}
		\footnotesize
		The best result (\textbf{bold}) is according to a two-valued t-test at a $5\%$ significance level.
	\end{center}
\end{table*}
\subsection{Effect of Link-Restore}
To investigate the effect of the proposed link-restore algorithm we apply it on 
GNLMF, RKNNLRS, NLSSC, and NLKSSC as a post-processing step.
This selection is based on the fact that link-restore is designed based on the non-negativity assumption about columns of $\nX$. Also regarding its application on GNLMF and NLSSC, we use the kernel matrix $\K(\nY,\nY)$ related to the kernel baselines. 
According to Table \ref{tab:link}, the application of link-restore was effecting regarding almost all the cases. It reduced the clustering error of all the relevant methods to some extent, which demonstrates its ability to correct broken links in the representation graph $\MC{G}$.
Nevertheless, the amount of improvements in NLSS/NLKSSC methods vary among datasets. 
For the 2-motions subset of Hopkins and for COIL20 datasets it did not add any important link to graph $\MC{G}$ which consequently did not change the value of $CE$. However, for YALE and AR datasets the amount of decreases in $CE$ shows the effectiveness of link-restore in correcting the missing connections in $\MC{G}$. 
\begin{figure}[tb]
	\centering
	\begin{subfigure}{0.45\textwidth}
		\centering		
		\includegraphics[width=.9\linewidth]{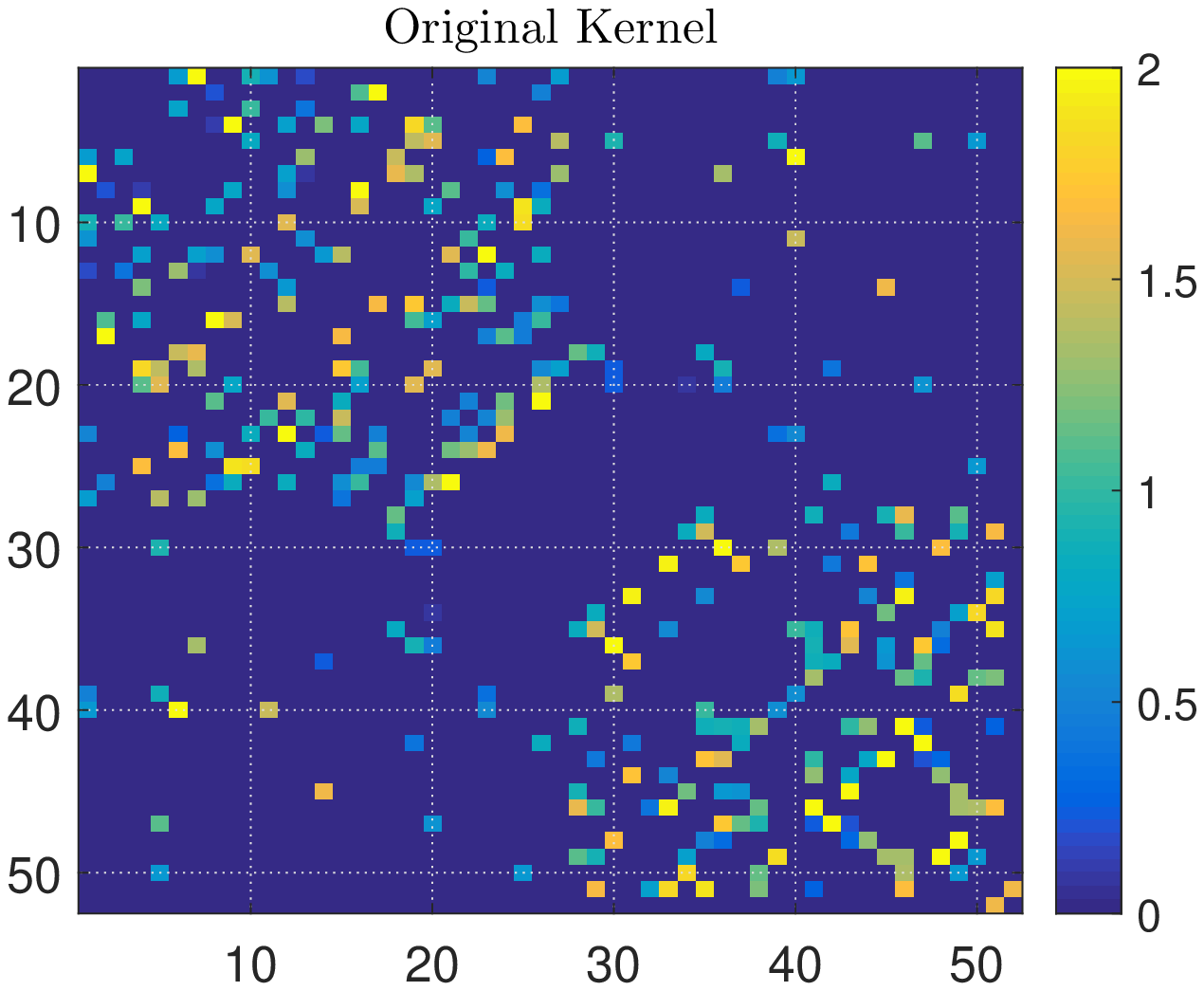}
		\caption{}
	\end{subfigure}	
	\begin{subfigure}{0.45\textwidth}
		\centering		
		\includegraphics[width=.9\linewidth]{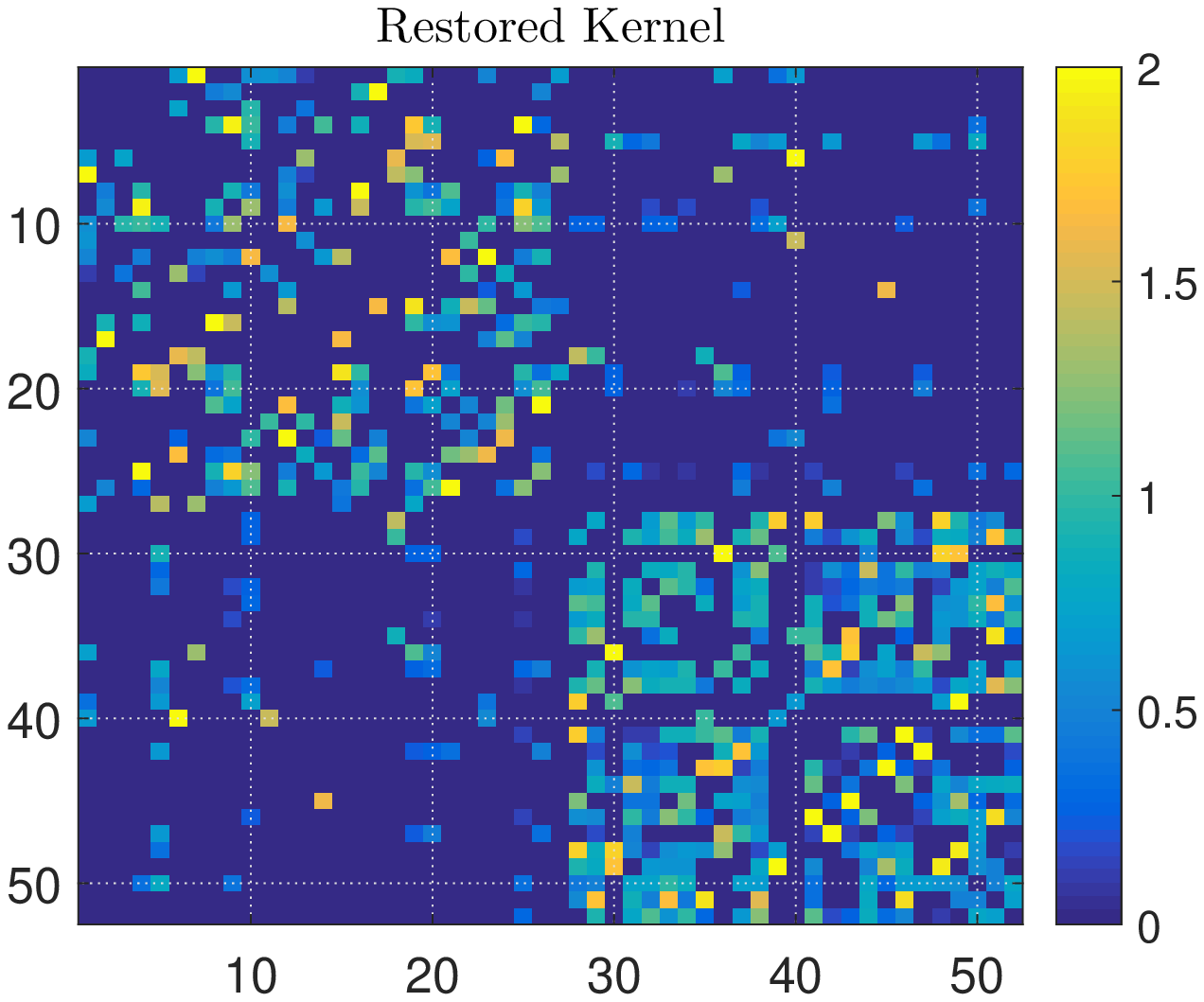}
		\caption{}			
	\end{subfigure}	
	\caption{A subset of the affinity matrix resulted by the implementation of NLKSSC on the AR dataset: (a) Before application of link-restore. (b) After application of link-restore.}
	\label{fig:link}
\end{figure}
Figure \ref{fig:link} visualizes the affinity matrix for implementation of NKLSSC on the AR dataset. The figure is zoomed in on two of the clusters showing that the representation graph contains more intra-cluster connections after applying link-restore (figure \ref{fig:link}-b). 
\subsection{Sensitivity to the Parameter Settings}
\begin{figure}[!b]
	\begin{subfigure}{0.32\textwidth}
		\centering		
		\includegraphics[width=.95\linewidth]{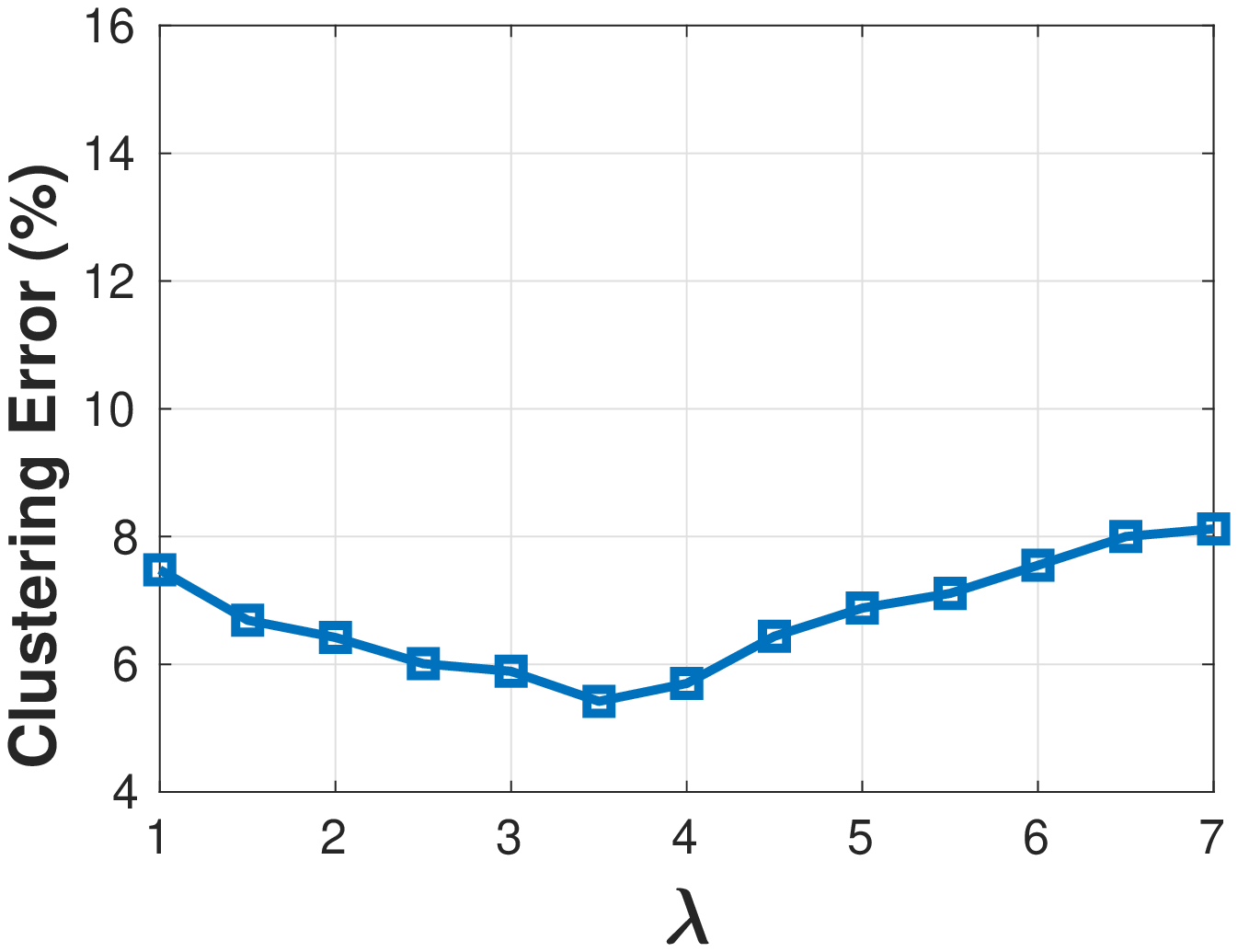}
		\caption{}
	\end{subfigure}	
	\begin{subfigure}{0.32\textwidth}
		\centering		
		\includegraphics[width=.95\linewidth]{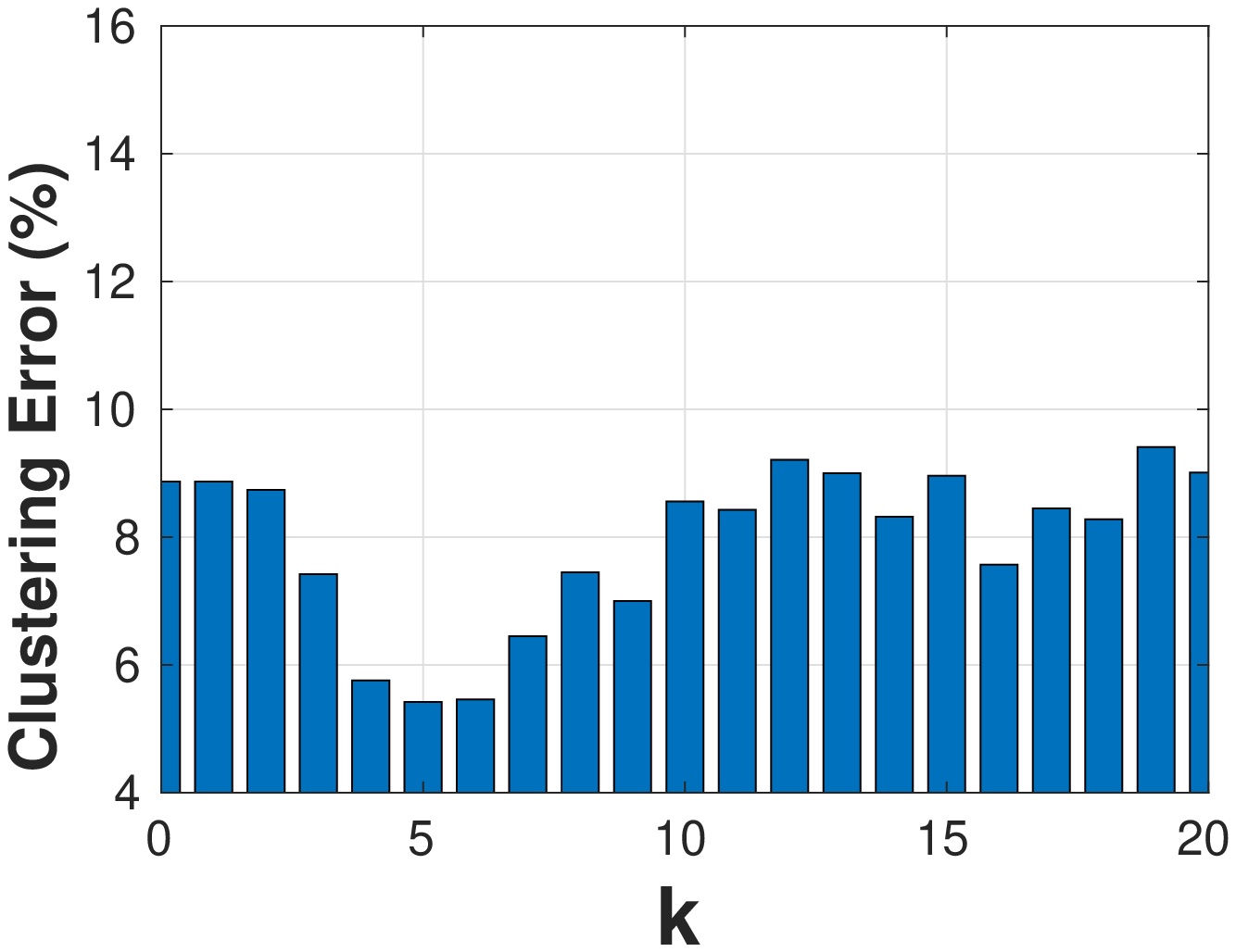}
		\caption{}			
	\end{subfigure}	
	\begin{subfigure}{0.32\textwidth}
		\centering		
		\includegraphics[width=0.95\linewidth]{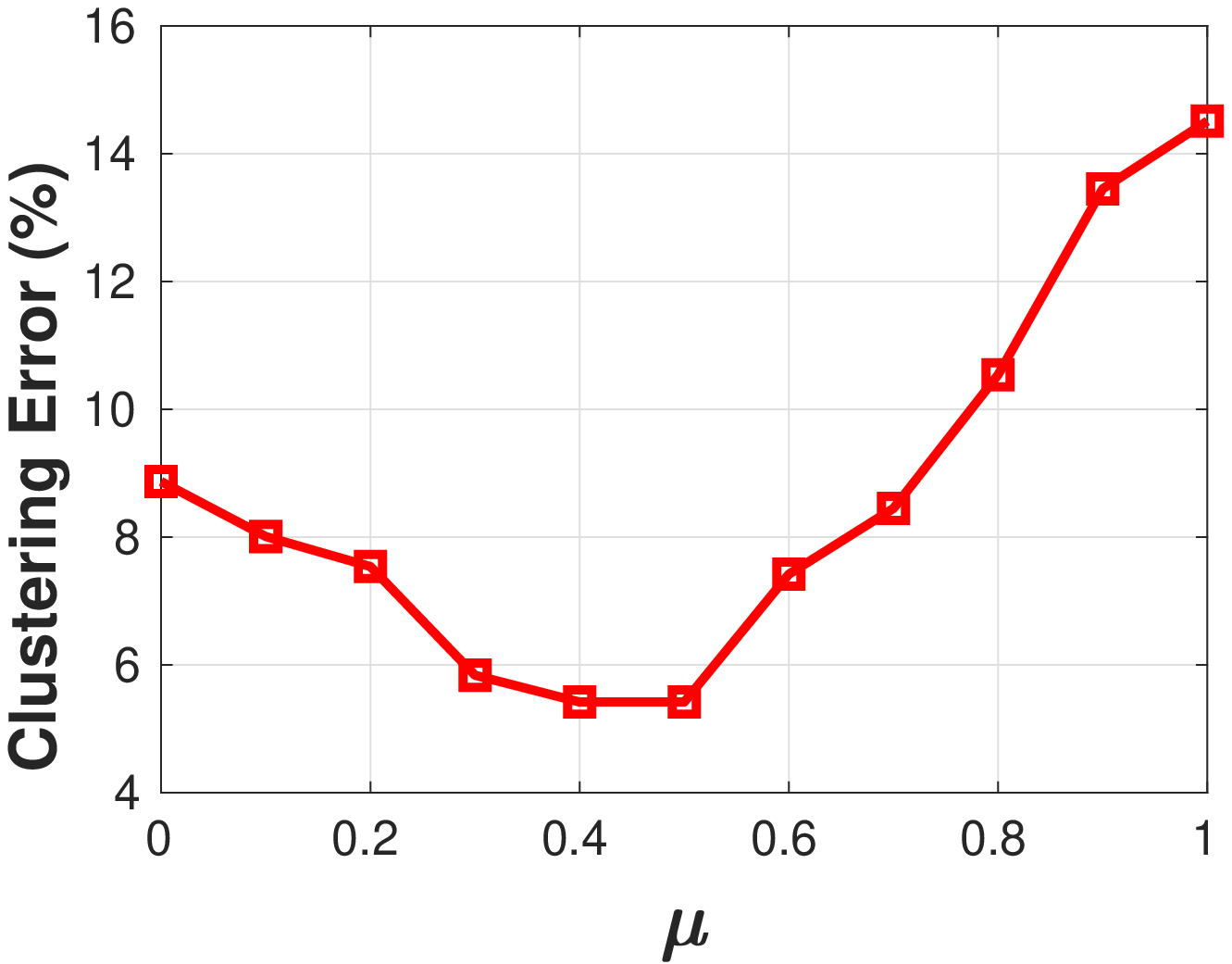}
		\caption{}		
	\end{subfigure}			
	\caption{Sensitivity analysis of NLKSSC to parameter selection (a)$\lambda$, (b)$\mu$ and (c)$k$ for AR dataset.}
	\label{fig:sens}
\end{figure}

Due to the space limits, we study the sensitivity of NLKSSC to the choice of parameters only for the AR dataset considering 3 different experiments. In each experiment, we fix two of $\lambda,\mu,k$ and change the other one and study the effect of this variation on clustering error ($CE$).
Based on Figure \ref{fig:sens}, the algorithm sensitivity to $\lambda$ is acceptable when $2\le\lambda\le4.5$. 
Having $\lambda \ge 6$ does not change $CE$ since it makes the loss term 
$\MC{E}_{rep}:=\| \Phi(\nY)-\Phi(\nY)\nX\|_F^2$ more dominant in optimization problem of Eq. \ref{eq:nklssc}.

By choosing $0.25 \le \mu \le 0.5$, the algorithm's performance does not change drastically. However, NLKSSC shows a considerable sensitivity if $\mu$ goes beyond 0.6. High values of $\mu$ weaken the role of $\MC{E}_{rep}$ (the main loss term) in the sparse coding model.

Studying the sensitivity curve of $k$, its starting point has a similar $CE$ to the start of $\mu$ sensitivity curve, as in both cases effect of $\MC{E}_{lsp}$ becomes zero in the optimization. Figure \ref{fig:sens}-b shows that $k\in\{3,4,5\}$ is a good choice. However, with $k\leq3$ the objective term $\MC{E}_{lsp}$ is not effective enough and with $k \ge 10$ the $CE$ curve does not follow any constant pattern, but generally becomes worse because it increases $\frac{\|W_w\|_0}{\|W_c\|_0}$ and it may infringe the pre-assumption of Proposition \ref{prop1}.
It is important to note that even a small neighborhood radius (e.g. $k=4$) could have a wide impact on the global representation if the local neighborhoods can have overlapping. Generally, similar sensitivity behaviors are also observed for the other datasets.
\section{Conclusion}
In this work, we proposed a novel subspace sparse coding framework regarding data clustering. 
Our non-negative local subspace clustering (NLSSC) benefits from a novel locality objective in its formulation which focuses on improving the separability of data points in the coding space.
In addition, NLSSC also obtains low-rank and affine sparse codes for the representation of the data. 
Implementations on real clustering benchmarks showed that this locality constraint is effective when performing a clustering based on the obtained representation graph.
In addition, the kernel extension of the algorithm (NLKSSC) is also provided in order to benefit from kernel-based representations of data.
Furthermore, we introduced the link-restore algorithm as an effective solution to the sparse coding redundancy issue when it has negative effects on clustering performance. This post-processing algorithm which is suitable for non-negative sparse representations corrects the broken links between close data points in the representation graph. 
Empirical evaluations demonstrated that link-restore can act as an effective post-processing step for different types of SSC methods which use non-negative sparse coding models.
As a future step, we are interested in combining our framework with dimension reduction strategies to better deal with multi-dimensional data types. 

\section{Acknowledgment}
This research was supported by the Cluster of Excellence Cognitive 
Interaction Technology 'CITEC' (EXC 277) at Bielefeld University, which
is funded by the German Research Foundation (DFG).
\bibliographystyle{IEEEtran}
\bibliography{/vol/semanticma/Thesis/Publications/Ref4Papers_CS}

\end{document}